\documentclass[letterpaper]{article} 
\usepackage[]{aaai25}  
\usepackage{times}  
\usepackage{helvet}  
\usepackage{courier}  
\usepackage[hyphens]{url}  
\usepackage{graphicx} 
\usepackage{natbib}  
\usepackage{caption} 
\usepackage{algorithm}
\usepackage{algorithmic}

\usepackage{newfloat}
\usepackage{listings}
\DeclareCaptionStyle{ruled}{labelfont=normalfont,labelsep=colon,strut=off} 
\floatstyle{ruled}
\newfloat{listing}{tb}{lst}{}
\floatname{listing}{Listing}
\usepackage{amsmath,amssymb,amsfonts}
\usepackage{multirow}

\title{CGLearn: Consistent Gradient-Based Learning for Out-of-Distribution Generalization}

\author {
    Jawad Chowdhury\textsuperscript{\rm 1},
    Gabriel Terejanu\textsuperscript{\rm 2}
}
\affiliations {
    Department of Computer Science, University of North Carolina at Charlotte, Charlotte, NC, USA \\
    mchowdh5@uncc.edu\textsuperscript{\rm 1}, 
    gabriel.terejanu@uncc.edu\textsuperscript{\rm 2} 
}

\usepackage{bibentry}

\begin{document}

\maketitle

\begin{abstract}
Improving generalization and achieving highly predictive, robust machine learning models necessitates learning the underlying causal structure of the variables of interest. A prominent and effective method for this is learning invariant predictors across multiple environments. In this work, we introduce a simple yet powerful approach, CGLearn, which relies on the agreement of gradients across various environments. This agreement serves as a powerful indication of reliable features, while disagreement suggests less reliability due to potential differences in underlying causal mechanisms. Our proposed method demonstrates superior performance compared to state-of-the-art methods in both linear and nonlinear settings across various regression and classification tasks. CGLearn shows robust applicability even in the absence of separate environments by exploiting invariance across different subsamples of observational data. Comprehensive experiments on both synthetic and real-world datasets highlight its effectiveness in diverse scenarios. Our findings underscore the importance of leveraging gradient agreement for learning causal invariance, providing a significant step forward in the field of robust machine learning. The source code of the linear and nonlinear implementation of CGLearn is open-source and available at:~\url{https://github.com/hasanjawad001/CGLearn}.
\end{abstract}
%
%
%
%
%
%
%
\section{Introduction}\label{sec:int}
Machine learning models have achieved remarkable success in various domains driven by the recent availability of large datasets, sophisticated algorithms, and highly advanced complex models. However, these models perform well only when the test data follows the same distribution as the training data (i.i.d.), but they often suffer from overfitting due to overparametrization, learning spurious correlations from training data~\cite{sagawa2020investigation, wang2021identifying, ming2022impact}. This issue arises because traditional models focus on predictive power without considering the causal relationships underlying the data. As a result, when the training and test distributions differ, models that rely on spurious correlations can perform very poorly, compromising their robustness, leading to poor generalization on out-of-distribution (OOD) test data~\cite{arjovsky2019invariant, he2021towards}.

\begin{figure*}[t]
  \centering
  \includegraphics[width=0.95\textwidth]{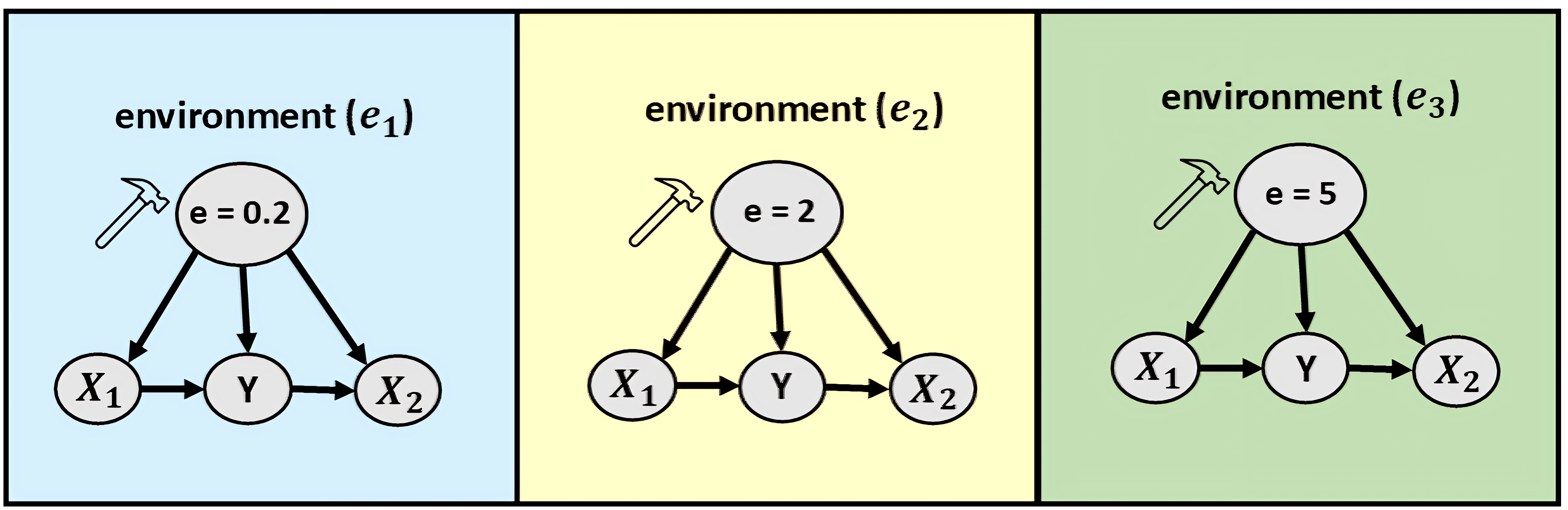}
  \caption{Illustration of three environments generated by intervening on the variable \(e\), which takes distinct values \(e = 0.2\), \(e = 2\), and \(e = 5\) in environments \(e_1\), \(e_2\), and \(e_3\), respectively. In each environment, \(X_1\) acts as a causal factor for the target variable \(Y\), while \(X_2\) is a spurious (non-causal) factor with respect to \(Y\). This figure exemplifies how different interventions on \(e\) create distinct environments.}
  \label{fig:mtd_env}
\end{figure*}
Learning causal relationships is the key to model explainability and enhancing generalization and robustness~\cite{shin2021effects, wang2022generalizing, santillan2023step}. Although the ideal method for learning causal structures is through Randomized Control Trials (RCTs), these are often expensive, unethical, or impractical. Various methods have been developed for causal discovery. Constraint-based methods use conditional independence tests to identify causal directions~\cite{spirtes2001causation, pearl2009causality, colombo2012learning}. This however often results in the Markov Equivalence Class (MEC) of causal structures. Score-based methods optimize causal graphs over Directed Acyclic Graphs (DAGs)~\cite{chickering2002optimal, ramsey2017million, huang2018generalized}, but the combinatorial nature of the search space can make it computationally expensive. Advances like NOTEARS~\cite{zheng2018dags} transform this combinatorial challenge into continuous optimization, leading to various effective variants~\cite{zheng2020learning, yu2019dag, lachapelle2019gradient, wei2020dags, ng2020role, ng2022masked}. However, learning causal structures purely from observational data can be challenging due to issues like selection bias, measurement errors, and confounding factors~\cite{zadrozny2004learning, torralba2011unbiased}. Moreover, relying solely on empirical risk optimization can result in models highly dependent on spurious relationships. To tackle this problem, researchers often use prior domain knowledge to improve causal discovery~\cite{o2006causal, gencoglu2020causal, andrews2020completeness, liu2021knowledge, chowdhury2023evaluation, chowdhury2023cd}. Unfortunately, many causal discovery methods depend on specific assumptions (e.g., linearity, non-Gaussian noise) that do not always hold in real-world data. In addition to that some of these methods exploit variance scales e.g. var-sortability to identify causal orderings, performing well on unstandardized data but poorly after standardization~\cite{reisach2021beware, kaiser2022unsuitability, reisach2024scale, ormaniec2024standardizing}. 
A recent line of study focuses on exploiting the invariance property of causal relationships across different environments. Methods like Invariant Causal Prediction (ICP)~\cite{peters2016causal} aim to identify causal predictors by ensuring the conditional distribution of the target given these predictors remains stable across environments. This method leverages the invariance of causal relationships under different interventions, iterating over feature subsets to find those invariant across environments, considering them as potential causal parents of the target variable. Another study, IRM~\cite{arjovsky2019invariant} optimizes a penalty function to achieve OOD generalization for predictive models, ensuring robust performance across environments. These methods significantly reduce the absorption of spurious correlations by focusing on stable and invariant relationships. The invariant learning framework provides a promising approach to improve model robustness and generalization in the presence of distribution shifts, with various domains exploiting invariance to learn better predictors and robust models~\cite{montavon2012learning, wang2017multiscale, chowdhury2024invariant, bose2024invariance}. Some relevant works such as AND-mask~\cite{parascandolo2020learning}, Fishr~\cite{rame2022fishr}, Fish~\cite{shi2021gradient}, IGA~\cite{koyama2020out} use environment-specific gradients to improve generalization in diverse settings. Moreover, approaches examining the signal-to-noise ratio (GSNR) in gradients, such as the work by Liu et al.~\cite{liu2020understanding}, measure the alignment of gradient directions across samples, while a similar strategy has been employed in large-batch training scenarios to improve model stability~\cite{jiang2023accelerating}.

Motivated by this line of work and the current drawbacks of existing methods in structure learning and OOD generalization, we introduce CGLearn, a general framework designed to improve the generalization of machine learning models by leveraging gradient consistency across different environments. CGLearn does not require extensive domain knowledge or assumptions over data linearity or noise, making it a versatile and practical approach for learning robust predictive models. By focusing on feature invariance, emphasizing on reliable features, and reducing dependence on spurious correlations, CGLearn enhances the reliability and robustness of the models. The main contributions of this study are stated as follows:
\begin{itemize}
    \item We propose a novel general framework, CGLearn, which improves consistency in learning robust predictors by focusing on features that show consistent behavior across environments.
    \item We provide both linear and nonlinear implementations of CGLearn, demonstrating its versatility and applicability across different model architectures.
    \item We demonstrate that CGLearn achieves superior predictive power and generalization, even without multiple environments, unlike most state-of-the-art methods in this arena that require diverse environments for effective generalization.
    \item Our empirical evaluations on synthetic and real-world datasets, covering both linear and nonlinear settings, as well as regression and classification tasks, validate the effectiveness and robustness of the proposed method.
\end{itemize}

The remainder of this paper is organized as follows: First, we delve into the methodology of CGLearn, detailing its linear and nonlinear implementations. Next, we present our experimental settings and evaluations. Finally, we encapsulate our conclusions, highlight the significant takeaways, and discuss future directions.

%
%
\section{Methodology}
\label{sec:mtd}
In this section, we present the methodology of CGLearn, detailing both its linear and nonlinear implementations. We start by explaining the regular Empirical Risk Minimization (ERM) approach and then introduce the concept of gradient consistency used in CGLearn. The primary concept of CGLearn is to enforce gradient consistency for each factor of our variable of interest across multiple environments to identify and utilize invariant features, thereby enhancing generalization and reducing dependence on spurious correlations.

\subsection{Empirical Risk Minimization (ERM)}
\label{sec:mtd_erm}
Let's consider a simple linear problem where the goal is to predict the target variable $Y$ using two features $X_1$ (causal) and $X_2$ (spurious) across multiple environments. Let $e_1, e_2, \ldots, e_m$ represent different environments. Environments can be considered as distinct distributions generated by different interventions, all of which share similar underlying causal mechanisms (see Fig.~\ref{fig:mtd_env}).

In the ERM framework, the weights for the features are updated by minimizing the empirical risk or the cost function ($L$), which is typically the mean squared error (MSE) between the predicted and actual values for a regression problem and cross-entropy loss for a classification task. Suppose the weights for the features at step $t$ are $w_1^t$ for $X_1$ and $w_2^t$ for $X_2$. The gradient of the loss ($L$) with respect to the weight associated with the \(j\)-th feature \(X_j\) in environment \(e_i\) is given by \(\nabla L_j^{e_i}\), where \(j \in \{1, 2\}\) and \(i \in \{1, \ldots, m\}\).

The aggregated gradient across all environments can be calculated as the mean of the gradients:
\begin{equation} \label{eqn:mtd_erm_mu}
\mu_j^{\text{grad}} = \frac{1}{m} \sum_{i=1}^{m} \nabla L_j^{e_i} \quad \text{for} \quad j \in \{1, 2\}
\end{equation}

Using this aggregated gradient, the weights are updated as follows:
\begin{equation} \label{eqn:mtd_erm_w}
w_j^{t+1} = w_j^t - \eta \mu_j^{\text{grad}} \quad \text{for} \quad j \in \{1, 2\}
\end{equation}
where $\eta$ is the learning rate. In this setup of a standard Empirical Risk Minimization, the weights for both $X_1$ and $X_2$ get updated in each step regardless of their consistency across environments. 

\begin{figure}[t]
  \centering
  \includegraphics[width=0.95\columnwidth]{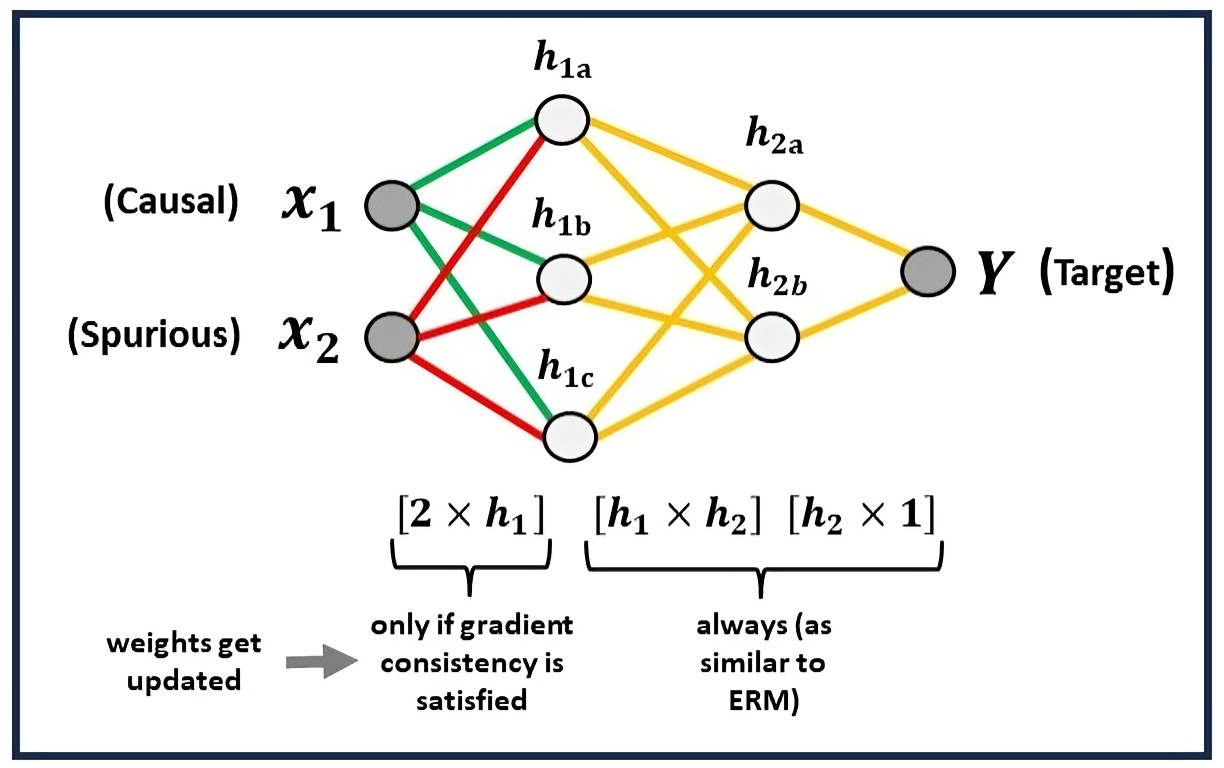}
  \caption{Nonlinear MLP implementation of CGLearn. \(X_1\) (causal) and \(X_2\) (spurious) feed into the first hidden layer $h_1$. Weight updates in $h_1$ are performed based on gradient consistency (using $L^2$-norm) for each feature across all training environments. The rest of the weights such as weights in $h_2$, are updated similarly to ERM (without imposing any consistency constraints).}
  \label{fig:exp_nme_cglearnMLP}
\end{figure}
\subsection{Linear Implementation of CGLearn}
\label{sec:mtd_lcg}

CGLearn modifies this approach by introducing a consistency check for the gradients. The idea is to update the weights only if the gradients are consistent across the available environments. This strategy focuses on invariant features and ignores spurious ones, expecting better generalization.

First, we calculate the gradient of each feature in every environment, as described in the previous section.
The mean of the gradients can be calculated as described in Eq.~\ref{eqn:mtd_erm_mu}. Next, we compute the standard deviation of the gradients for each feature across all environments as follows:
\begin{equation} \label{eqn:mtd_lcg_sig}
\sigma_j^{\text{grad}} = \sqrt{\frac{1}{m} \sum_{i=1}^{m} \left( \nabla L_j^{e_i} - \mu_j^{\text{grad}} \right)^2}
\end{equation}

We then calculate the consistency ratio, which is the absolute value of the ratio of the mean gradient to the standard deviation of the gradients:
\begin{equation} \label{eqn:mtd_lcg_cr}
C_j^{\text{ratio}} = \left| \frac{\mu_j^{\text{grad}}}{\sigma_j^{\text{grad}}} \right|
\end{equation}

The consistency ratio, $C_j^{\text{ratio}}$ defined in Eq.~\ref{eqn:mtd_lcg_cr}, is considered to be an indicator of the invariance of the gradient of variable $X_j$ across all the training environments. A relatively larger mean compared to the standard deviation would indicate more similar or invariant gradients across the environments for the feature $X_j$, resulting in a higher value of $C_j^{\text{ratio}}$. On the other hand, a larger standard deviation indicates more diversity across the environments for $X_j$. Finally, we formulate a consistency mask based on a predefined threshold $C^{\text{thresh}}$:
\begin{equation} \label{eqn:mtd_lcg_cm}
C_j^{\text{mask}} = 
\begin{cases} 
1 & \text{if } C_j^{\text{ratio}} \geq C^{\text{thresh}} \\
0 & \text{otherwise}
\end{cases}
\end{equation}

The weights are updated only for the feature that has a nonzero mask and remains unchanged otherwise as per the following equation:
\begin{equation} \label{eqn:mtd_lcg_w}
w_j^{t+1} = w_j^t - \eta \left( \mu_j^{\text{grad}} \cdot C_j^{\text{mask}} \right) \quad \text{for} \quad j \in \{1, 2\}
\end{equation}

Considering our motivating example, where \(X_1\) is causal and expected to show more consistency across environments, \(C_1^{\text{mask}}\) is expected to be 1. Conversely, \(X_2\) is spurious with respect to the target, expected to show inconsistency across environments, and \(C_2^{\text{mask}}\) is expected to be 0. Therefore, the weight for $X_1$ is mostly updated throughout the training steps while the weight for $X_2$ is not. The model thus focuses on the features that show consistency for learning the predictors of the target. This implementation strategy ensures to emphasis on reliable, invariant features while minimizing the impact of unreliable features by keeping their weights unchanged (or keeping the changes to a minimum). As a result, the contributions of the spurious features remain constant in the context of the model updates. In the next section, we extend the CGLearn method to a nonlinear setting using multilayer perceptron (MLP) as an instance.

\begin{figure*}[t]
  \centering
  \includegraphics[width=0.95\textwidth]{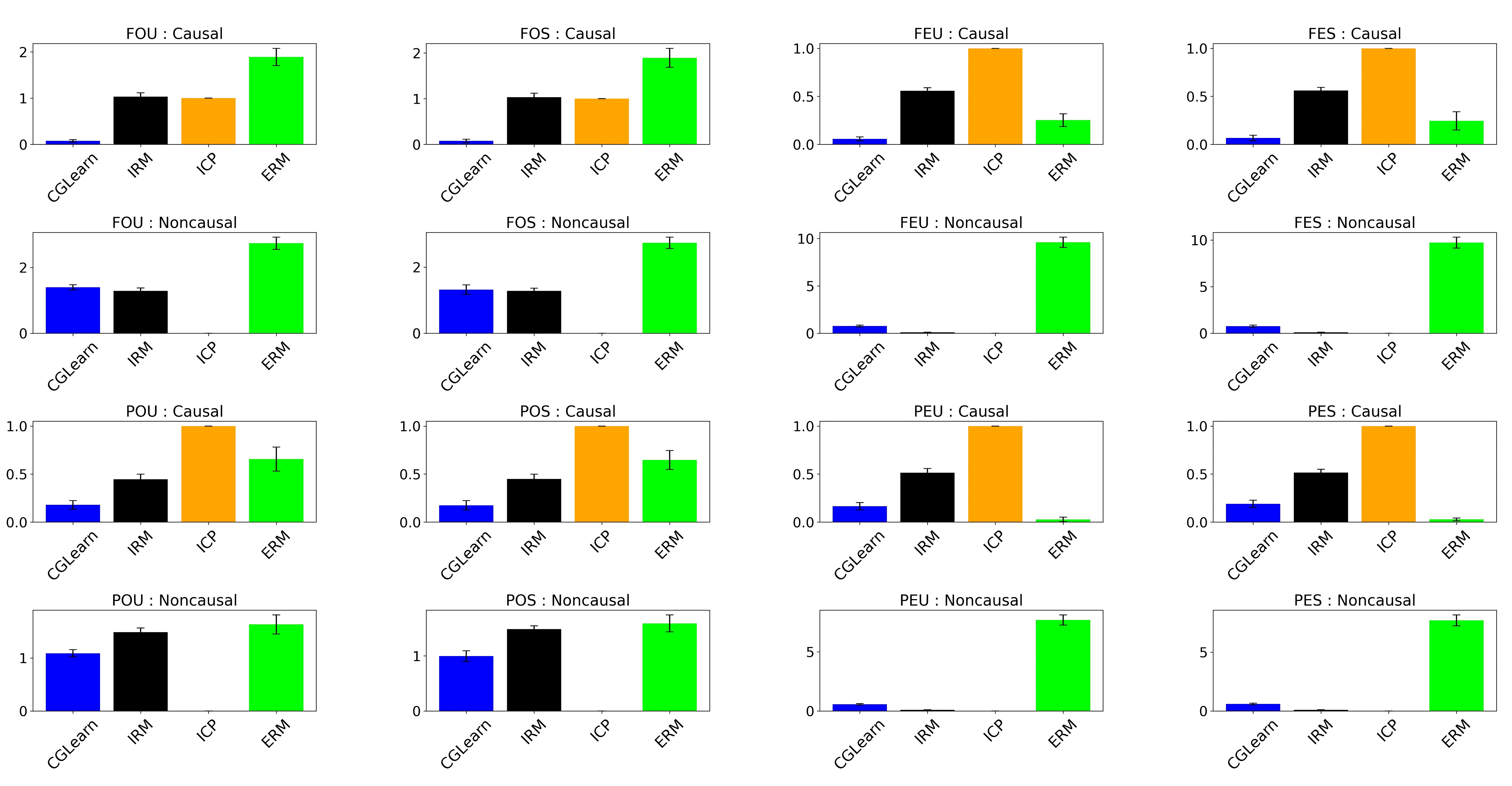}
  \caption{Performance comparison of CGLearn, IRM, ICP, and ERM across various linear multiple environment setups. Each subplot represents different configurations of the data, showing the mean squared error (MSE) for causal and noncausal variables over 50 trials.}
  \label{fig:exp_lme}
\end{figure*}
\subsection{Nonlinear Implementation}
\label{sec:mtd_nonlinear}

For the nonlinear implementation of CGLearn using a multilayer perceptron (MLP), we focus on the gradients in the first hidden layer ($h_1$), where feature contributions can be distinctly identified. By controlling the contribution of spurious features at the first hidden layer, we ensure they do not influence the final output. The process involves calculating the $L^2$-norm of the gradients for each feature in each environment, followed by determining the consistency ratio and mask to impose the consistency constraint.

${\|\nabla L}_{jh_1}^{e_i}\|_{2}$ denotes the $L^2$-norm of the gradients of the \( j \)-th feature $X_j$ in the \( i \)-th environment $e_i$ at the first hidden layer $h_1$. We compute the mean and standard deviation of the $L^2$-norm of the gradients across all environments as follows:
\begin{equation} \label{eqn:mtd_ncg_mu}
\mu_j^{\text{grad}} = \frac{1}{m} \sum_{i=1}^{m} {\|\nabla L}_{jh_1}^{e_i}\|_{2}
\end{equation}
\begin{equation} \label{eqn:mtd_ncg_sig}
\sigma_j^{\text{grad}} = \sqrt{\frac{1}{m} \sum_{i=1}^{m} \left( {\|\nabla L}_{jh_1}^{e_i}\|_{2} - \mu_j^{\text{grad}} \right)^2}
\end{equation}

We then calculate the consistency ratio, $C_j^{\text{ratio}}$ and the consistency mask, $C_j^{\text{mask}}$ for feature $X_j$ by following Eq.~\ref{eqn:mtd_lcg_cr} and~\ref{eqn:mtd_lcg_cm} respectively. All the weights that belong to a particular feature, $X_j$ in the first hidden layer $h_1$, are updated by following a similar strategy to Eq.~\ref{eqn:mtd_lcg_w}. This updating strategy that depends on the consistency ratio, ensures that only the features that show consistency across the environments are considered to be updated. Otherwise, the weights remain unchanged, effectively treating them as constants similar to the linear implementation. For weights corresponding to the rest of the model other than the first hidden layer are updated as similar to ERM. 

Fig.~\ref{fig:exp_nme_cglearnMLP} illustrates a simple demonstration of the nonlinear MLP implementation of CGLearn. In this figure, \(X_1\) and \(X_2\) represent causal and spurious features, respectively, in accordance with our earlier motivating example. The gradient consistency is checked in the first hidden layer (\(h_1\)), and weights are updated only if the consistency ratio exceeds the threshold, ensuring that features that show invariance across environments are utilized.

In both implementations, the goal is to ensure that the model relies on features that show invariance across different environments. This leads to more robust and generalizable models by reducing dependency on spurious correlations.

\section{Experiments and Results}
\label{sec:exp}
We have considered three different major scenarios to assess the predictivity, robustness, and generalization capabilities of CGLearn. The first two scenarios are the ones where we considered linearly generated dataset-based experiments and in the last experimental case we have used the nonlinear implementation of CGLearn using multilayer perceptron (MLP) and applied it to different real world regression and classification tasks. 

For all evaluations, we reported the mean and standard deviation of the performance metrics considered. For statistical significance tests, we used a t-test with $\alpha = 0.05$ as the significance level.

\begin{table*}[t]
    \centering
    \begin{tabular}{|l|ll|ll|}
    \hline
    \multirow{2}{*}{Cases} & \multicolumn{2}{l|}{Causal Error (MSE)}                 & \multicolumn{2}{l|}{Noncausal Error (MSE)}                      \\ \cline{2-5} 
                           & \multicolumn{1}{l|}{CGLearn}              & ERM         & \multicolumn{1}{l|}{CGLearn}              & ERM                  \\ \hline
    FOU                    & \multicolumn{1}{l|}{\textbf{1.28 ± 0.40}} & 1.57 ± 0.13 & \multicolumn{1}{l|}{0.61 ± 0.19}          & \textbf{0.54 ± 0.05} \\ \hline
    FOS                    & \multicolumn{1}{l|}{\textbf{1.40 ± 0.43}} & 1.61 ± 0.10 & \multicolumn{1}{l|}{0.53 ± 0.17}          & 0.52 ± 0.06          \\ \hline
    FEU                    & \multicolumn{1}{l|}{\textbf{0.13 ± 0.05}} & 0.20 ± 0.04 & \multicolumn{1}{l|}{\textbf{7.22 ± 2.15}} & 8.28 ± 0.28          \\ \hline
    FES                    & \multicolumn{1}{l|}{\textbf{0.16 ± 0.06}} & 0.20 ± 0.04 & \multicolumn{1}{l|}{\textbf{7.47 ± 2.23}} & 8.36 ± 0.30          \\ \hline
    POU                    & \multicolumn{1}{l|}{\textbf{0.28 ± 0.11}} & 0.37 ± 0.08 & \multicolumn{1}{l|}{0.51 ± 0.18}          & 0.48 ± 0.11          \\ \hline
    POS                    & \multicolumn{1}{l|}{\textbf{0.34 ± 0.13}} & 0.39 ± 0.07 & \multicolumn{1}{l|}{0.46 ± 0.17}          & 0.48 ± 0.10          \\ \hline
    PEU                    & \multicolumn{1}{l|}{\textbf{0.24 ± 0.10}} & 0.32 ± 0.07 & \multicolumn{1}{l|}{\textbf{5.11 ± 1.57}} & 5.83 ± 0.43          \\ \hline
    PES                    & \multicolumn{1}{l|}{\textbf{0.26 ± 0.10}} & 0.31 ± 0.06 & \multicolumn{1}{l|}{\textbf{5.21 ± 1.58}} & 5.81 ± 0.36          \\ \hline
    \end{tabular}
    \caption{Performance evaluation of CGLearn and ERM in linear single environmental setups. The table shows the Mean Squared Errors (MSE) for causal and noncausal variables across 50 trials for each configuration. Bold values indicate statistical significance.}
    \label{tab:exp_lse}
\end{table*}
\subsection{Linear Multiple Environments}
\label{sec:exp_lme}
To evaluate the performance of our proposed CGLearn method, we generated synthetic linear datasets inspired by the approach used in the Invariant Risk Minimization (IRM) framework \cite{arjovsky2019invariant}. Our goal was to create diverse environments to test the robustness of our model under varying conditions.

We generated eight different experimental setups based on three key factors. Each setup included datasets with one target variable \(Y\) and ten feature variables \(X_1\) to \(X_{10}\). Features \(X_1\) to \(X_5\) acted as causal parents of \(Y\), while \(X_6\) to \(X_{10}\) were influenced by \(Y\) (non-causal). First, we distinguished between scrambled (S) and unscrambled (U) observations by applying an orthogonal transformation matrix \(S\) for scrambled data and using the identity matrix \(I\) for unscrambled data. This scrambling ensures that the features are not directly aligned with their original scales, making the learning task more challenging. Second, we designed fully-observed (F) scenarios where hidden confounders did not directly affect the features (i.e., no hidden confounder effects on features), and partially-observed (P) scenarios where hidden confounders influenced the features with Gaussian noise. Third, we incorporated two types of noise for the target variable \(Y\): homoskedastic (O) noise, where the noise variance remained constant across different environments, and heteroskedastic (E) noise, where the noise variance varied depending on the environment, increasing with higher values of \(e\). This distinction captures different real-world scenarios where noise may or may not depend on external factors. For each of these eight configurations (combinations of S/U, F/P, and O/E), we generated datasets corresponding to three distinct environments defined by the values \(e \in \{0.2, 2, 5\}\). Each dataset consisted of 1000 samples. To ensure consistency with the IRM methodology and experimental setup, we used \(e = 5\) as the validation environment and determined the optimal consistency threshold \((C^{\text{thresh}})\) for our CGLearn method using the performance based on this validation data. We selected the threshold \(C^{\text{thresh}}\) from the candidate values \(\{0.25, 1, 4, 16, 64\}\) based on validation performance. This threshold is critical for identifying the invariant and most reliable features across different environments. For more details on the data generation process, we refer readers to the IRM paper \cite{arjovsky2019invariant}.

We compared the performance of CGLearn with Empirical Risk Minimization (ERM), Invariant Causal Prediction (ICP) \cite{peters2016causal}, and IRM \cite{arjovsky2019invariant}. We considered $50$ random trials and reported the results in Fig.~\ref{fig:exp_lme}. In most cases, our proposed method CGLearn achieves the lowest mean squared error (MSE), demonstrating superior performance across various test cases to distinguish the causal and noncausal factors of the target by exploiting invariance across environments. IRM performs better than ERM but does not match the accuracy of CGLearn. ERM shows the highest errors in most cases, as it fails to differentiate between causal and noncausal features, relying on spurious correlations. Interestingly, ICP performs well in noncausal scenarios but poorly in causal ones. This observation aligns with the findings from the IRM study \cite{arjovsky2019invariant}, which noted that ICP's conservative nature leads it to reject most covariates as direct causes, resulting in high causal errors.

\subsection{Linear Single Environment}
\label{sec:exp_lse}
To evaluate the performance of our proposed CGLearn method in scenarios with only one environment, we generated synthetic linear datasets without relying on multiple environments as in previous experiments. For each of the eight cases, we used a single setting with \(e = 2\). The data generation process was similar to the previous section, with each dataset consisting of 1000 samples and ten feature variables, \(X_1\) to \(X_{10}\). The first five features (\(X_1\) to \(X_5\)) acted as causal parents of the target variable \(Y\), while the remaining five features (\(X_6\) to \(X_{10}\)) were influenced by \(Y\). Given the single environment setup, we could not apply IRM and ICP methods, as they require multiple environments to distinguish between causal and noncausal factors. Therefore, we compared our results solely with Empirical Risk Minimization (ERM).

In the case of CGLearn, we created multiple batches, with \(b = \{3, 5\}\) representing the number of batches created from the dataset. The last batch was used as the validation batch to determine the optimal consistency threshold parameter \((C^{\text{thresh}})\). We selected the threshold \(C^{\text{thresh}}\) from the candidate values \(\{0.25, 1, 4, 16, 64\}\) based on validation performance. We imposed gradient consistency across different batches to learn consistent and reliable factors of the target.

Table \ref{tab:exp_lse} shows the results of our experiments in the single environment setup. Considering the causal error across all eight cases, CGLearn consistently achieves significantly lower mean squared errors (MSE) compared to ERM. For the noncausal error, CGLearn also outperforms ERM in most cases, suggesting the superiority of the proposed approach. Even in the absence of multiple environments, the optimization strategy based on gradient consistency across different batches enables CGLearn to achieve better predictive power than standard ERM.

\begin{table*}[t]
    \centering
    \begin{tabular}{|l|c|l|l|l|}
    \hline
    Dataset                 & \# Optimal Envs.   & Method  & RMSE (Train) & RMSE (Test)          \\ \hline
    \multirow{4}{*}{Boston} & \multirow{4}{*}{7} & ERM     & 3.57 ± 0.11  & 6.43 ± 0.45           \\ \cline{3-5} 
                            &                    & IRM     & 3.79 ± 0.33  & 6.99 ± 0.74           \\ \cline{3-5} 
                            &                    & BIRM    & 3.77 ± 0.50  & 7.70 ± 0.52           \\ \cline{3-5}
                            &                    & CGLearn & 1.91 ± 0.26  & \textbf{5.49 ± 0.28}  \\ \hline
    \multirow{4}{*}{Yacht}  & \multirow{4}{*}{5} & ERM     & 0.21 ± 0.04  & 3.47 ± 1.15           \\ \cline{3-5} 
                            &                    & IRM     & 2.90 ± 0.03  & 4.36 ± 0.38           \\ \cline{3-5} 
                            &                    & BIRM    & 0.71 ± 0.19  & 3.15 ± 0.75           \\ \cline{3-5}
                            &                    & CGLearn & 0.48 ± 0.23  & \textbf{2.29 ± 0.42}  \\ \hline
    \end{tabular}
    \caption{Performance comparison in nonlinear experimental setups for regression tasks. The table shows the RMSE for training and test environments across 10 trials. In test cases, we mark the statistically significant values in bold.}
    \label{tab:exp_nme_reg}
\end{table*}
\subsection{Nonlinear Multiple Environments} \label{sec:exp_nme}
For the nonlinear experimental setups, we considered two types of supervised learning tasks: regression and classification, both on real-world datasets. This approach allows us to evaluate the performance and robustness of our proposed CGLearn method in different real-world contexts. Recent work has highlighted limitations in the original Invariant Risk Minimization (IRM) framework, particularly in nonlinear settings where deep models tend to overfit \cite{rosenfeldrisks}. To address this, we included Bayesian Invariant Risk Minimization (BIRM) as a baseline, which has been shown to alleviate overfitting issues by incorporating Bayesian inference and thereby improving generalization in nonlinear scenarios \cite{lin2022bayesian}.

\textbf{Regression Tasks.} In the nonlinear implementation of CGLearn, we used a multilayer perceptron (MLP) to evaluate its performance on real-world regression tasks, comparing it with other baselines. For the regression tasks, we used the Boston Housing dataset \cite{harrison1978hedonic} and the Yacht Hydrodynamics dataset \cite{misc_yacht_hydrodynamics_243}.
The Boston Housing dataset consists of $506$ instances and $13$ continuous attributes. It concerns housing values in suburbs of Boston, with the task being to predict the median value of owner-occupied homes (MEDV) based on attributes such as per capita crime rate (CRIM), proportion of residential land zoned for large lots (ZN), average number of rooms per dwelling, and etc. The Yacht Hydrodynamics dataset consists of $308$ instances and $6$ attributes. The task is to predict the residuary resistance per unit weight of displacement of a yacht based on various hull geometry coefficients and the Froude number, such as the longitudinal position of the center of buoyancy, prismatic coefficient, and beam-draught ratio.

Since real-world datasets do not naturally come with different environments, we followed a similar approach to the study by Ge et al. \shortcite{ge2022invariant}. We used the K-Means~\cite{lloyd1982least} clustering algorithm to generate diverse environments and determined the optimal number of environments (between $3$ to $10$) using the Silhouette~\cite{rousseeuw1987silhouettes} method. For each dataset, we created all possible test cases where each environment was considered as the test environment once, and the rest were used as training environments. We averaged the results over all possible test cases and repeated the process for $10$ random trials. We evaluated the models based on RMSE, with the results shown in Table \ref{tab:exp_nme_reg}. For the Boston Housing dataset, we found the optimal number of environments was $7$, while for the Yacht Hydrodynamics dataset, it was $5$. From Table \ref{tab:exp_nme_reg}, we observe that all four methods perform better on the training environments than the test environments, as expected. However, CGLearn shows significantly lower error in the testing or unseen environments compared to the other methods, demonstrating that imposing gradient consistency leads to less dependence on spurious features and thus better generalization.

\textbf{Classification Tasks.} For the classification tasks, we evaluated the performance on two real-world classification datasets: the Wine Quality dataset for red and white wines from the UCI repository \cite{misc_wine_quality_186}. The Wine Quality dataset for red wine has $1599$ instances and $11$ attributes, while the dataset for white wine has $4898$ instances and $11$ attributes. The goal is to model wine quality based on physicochemical tests, such as fixed acidity, volatile acidity, citric acid, residual sugar, pH, and etc.
Similar to the regression tasks, we used K-means clustering to generate diverse environments and determined the optimal number of environments using the Silhouette method, finding $4$ as the optimal number of environments for both classification datasets. We then generated all possible test cases where each environment was considered the test environment once, and the rest were used as training environments (as we did with the regression tasks). We averaged the performance over all possible test cases and conducted the process for $10$ random trials. We used accuracy and F1-score as evaluation metrics, with the results shown in Table \ref{tab:exp_nme_cls}.
\begin{table*}[t]
    \centering
    \begin{tabular}{|l|c|l|l|l|l|l|}
    \hline
    Dataset                             & \# Optimal Envs.   & Method  & Accuracy (Train) & Accuracy (Test)       & F1-score (Train) & F1-score (Test)        \\ \hline
    \multirow{4}{*}{WQ Red}   & \multirow{4}{*}{4} & ERM     & 62.07 ± 0.34     & 58.08 ± 1.72          & 0.692 ± 0.004    & 0.535 ± 0.010          \\ \cline{3-7} 
                                        &                    & IRM     & 63.68 ± 0.19     & 58.70 ± 1.54          & 0.644 ± 0.003    & 0.542 ± 0.014          \\ \cline{3-7} 
                                        &                    & BIRM    & 64.94 ± 0.37     & 57.97 ± 0.93          & 0.626 ± 0.004    & 0.536 ± 0.011          \\ \cline{3-7}
                                        &                    & CGLearn & 61.59 ± 0.44     & 59.60 ± 0.46          & 0.638 ± 0.008    & \textbf{0.553 ± 0.007} \\ \hline
                                        
    \multirow{4}{*}{WQ White} & \multirow{4}{*}{4} & ERM     & 58.73 ± 0.22     & 51.15 ± 0.34          & 0.590 ± 0.002    & 0.447 ± 0.008          \\ \cline{3-7} 
                                        &                    & IRM     & 58.82 ± 0.26     & 51.60 ± 0.40          & 0.566 ± 0.003    & 0.450 ± 0.013          \\ \cline{3-7} 
                                        &                    & BIRM    & 58.04 ± 0.18     & 51.87 ± 0.32          & 0.530 ± 0.006    & 0.460 ± 0.026          \\ \cline{3-7}
                                        &                    & CGLearn & 58.23 ± 0.38     & \textbf{52.33 ± 0.32} & 0.555 ± 0.005    & 0.460 ± 0.007          \\ \hline
    
    \end{tabular}
    \caption{Performance comparison in nonlinear setups for classification tasks. The table shows accuracy and F1-score for training and test environments across 10 trials. The statistically significant values are in bold for the test cases. WQ Red and WQ White represent the Wine Quality Red and Wine Quality White datasets respectively.}
    \label{tab:exp_nme_cls}
\end{table*}
As expected, all methods performed better in training environments compared to test environments. However, we found that CGLearn achieved higher accuracy and F1-scores, which are desirable, and the superior performance was statistically significant for the F1-score on the Wine Quality Red dataset. It also had significantly better accuracy on the Wine Quality White dataset. Similar to the regression tasks, CGLearn demonstrated better predictive power and generalization over ERM, IRM, and BIRM for the classification tasks.

\textbf{Limitations of CGLearn with Invariant Spurious Features.} We evaluated CGLearn on the Colored MNIST dataset, a synthetic binary classification task derived from MNIST \cite{lecun1995learning} and proposed in the IRM study~\cite{arjovsky2019invariant}. This dataset introduces color as a spurious feature that strongly correlates with the label in the training environments but has the correlation reversed in the test environment. We applied the nonlinear implementation of CGLearn and compared it with the results of ERM and IRM as reported in the IRM study~\cite{arjovsky2019invariant}. Over 10 trials, ERM achieved a training accuracy of 87.4 ± 0.2 and a test accuracy of 17.1 ± 0.6, while IRM achieved a training accuracy of 70.8 ± 0.9 and a test accuracy of 66.9 ± 2.5. In our experimental study, CGLearn achieved a training accuracy of 93.1 ± 0.8 and a test accuracy of 29.1 ± 0.8. While CGLearn slightly outperformed ERM in the test environment, it still struggled to generalize. This limitation arises because CGLearn imposes gradient consistency on the training environments to distinguish invariant features from spurious ones. However, in the Colored MNIST setup, the spurious feature (color) is consistent across both training environments, leading CGLearn to erroneously treat it as an invariant feature. Consequently, CGLearn relies on color and performs poorly in the test environment. To improve CGLearn's generalization, future work should focus on adapting the method to account for the varying nature of spurious features, even when they appear consistent across training environments.
\section{Conclusions} \label{sec:con}
In this study, we presented CGLearn, a novel approach for developing robust and predictive machine learning models by leveraging gradient consistency across multiple environments. By focusing on the agreement of gradients, CGLearn effectively identifies and utilizes invariant features, leading to superior generalization and reduced reliance on spurious correlations. Our extensive experiments on both synthetic and real-world datasets, including regression and classification tasks, demonstrated that CGLearn outperforms traditional ERM and state-of-the-art invariant learners like ICP, IRM, and BIRM, achieving lower errors and better generalization in diverse scenarios. Notably, even in the absence of predefined environments, we demonstrated that CGLearn can be effectively applied to different subsamples of data, leading to better predictive models than regular ERM. This flexibility enhances the applicability of CGLearn in a wide range of real-world scenarios where many state-of-the-art methods require diverse and defined environments for OOD generalization.

Despite its strengths, CGLearn has limitations, particularly in scenarios where spurious features are invariant across environments, as observed in the Colored MNIST experiments. Such cases violate our assumption as generally we expect and observe causal features to be stable and invariant in nature whereas spurious features do not~\cite {woodward2005making, wang2022generalizing}. CGLearn erroneously considers these invariant but spurious features as reliable, impacting its generalization performance. Addressing this limitation and adapting CGLearn to better handle such cases is a promising direction for future research.

Overall, CGLearn provides a significant step forward in the field of robust machine learning by effectively harnessing causal invariance. Our work opens new avenues for developing models that are not only highly predictive but also resilient to distribution shifts, paving the way for more reliable applications in real-world settings.

\section{Acknowledgments}
This research was supported by the Army Research Office under Grant W911NF-22-1-0035 and by the National Science Foundation under grant no. CBET 2218841. The views expressed are those of the authors and do not necessarily reflect the official policies of the Army Research Office or the U.S. Government. The U.S. Government retains the right to reproduce and distribute reprints for governmental purposes.

%
%

\end{document}